# Mining GIS Data to Predict Urban Sprawl

Anita Pampoore-Thampi[1], Aparna S. Varde[1,3], Danlin Yu[2,3]
1. Department of Computer Science
2. Department of Earth and Environmental Studies
3. Environmental Management PhD Program
Montclair State University, Montclair, NJ, USA
(pampooretha1 | vardea | yud)@montclair.edu

## ABSTRACT
This paper addresses the interesting problem of processing and analyzing data in geographic information systems (GIS) to achieve a clear perspective on urban sprawl. The term "urban sprawl" refers to overgrowth and expansion of low-density areas with issues such as car dependency and segregation between residential versus commercial use. Sprawl has impacts on the environment and public health. In our work, spatiotemporal features related to real GIS data on urban sprawl such as population growth and demographics are mined to discover knowledge for decision support. We adapt data mining algorithms, Apriori for association rule mining and J4.8 for decision tree classification to geospatial analysis, deploying the ArcGIS tool for mapping. Knowledge discovered by mining this spatiotemporal data is used to implement a prototype spatial decision support system (SDSS). This SDSS predicts whether "urban sprawl" is likely to occur. Further, it estimates the values of pertinent variables to understand how the variables impact each other. The SDSS can help decision-makers identify problems and create solutions for avoiding future sprawl occurrence and conducting urban planning where sprawl already occurs, thus aiding sustainable development. This work falls in the broad realm of geospatial intelligence and sets the stage for designing a large scale SDSS to process big data in complex environments, which constitutes part of our future work.

## Categories and Subject Descriptors
H.2.8 [**Database Management**]: Database Applications – data mining, scientific databases, spatial databases and GIS; I.2.1 [**Artificial Intelligence**]: Applications and Expert Systems – industrial automation

## 1. INTRODUCTION
We are living in a world which grows and urbanizes at a rapid pace. If we continue this trend, our successors will be left with no resources to sustain. Other impacts of this growth that occur when mankind spreads its wings to the outskirts of the urban background are: overcrowding, pollution, unemployment, crime, poverty, disease etc. Today, expansion implies that cities are growing to nearby towns and villages by converting those natural lands to impervious lands by constructing buildings, parking lots, highways etc. Thus there should be some method to curb this urban overgrowth and impose some limitations for urbanization in each area. This can only be achieved by taking appropriate decisions. Urban planners and engineers should take appropriate decisions to protect natural land while designing any activities for new constructions. With these concepts, we introduce a prototype spatial decision support system (SDSS) by mining geospatial data on parameters relevant to urban planning. A decision support system falls in the category of expert systems designed to support or assist the users' decisions in specific domains, thus playing the decision-making role of an expert. SDSS refers to spatial aspects of the data, such as location-specific parameters, hence the term spatial decision support system [20]. This paper aims to formulate a model which predicts the occurrence of a concept known as *urban sprawl* explained next.

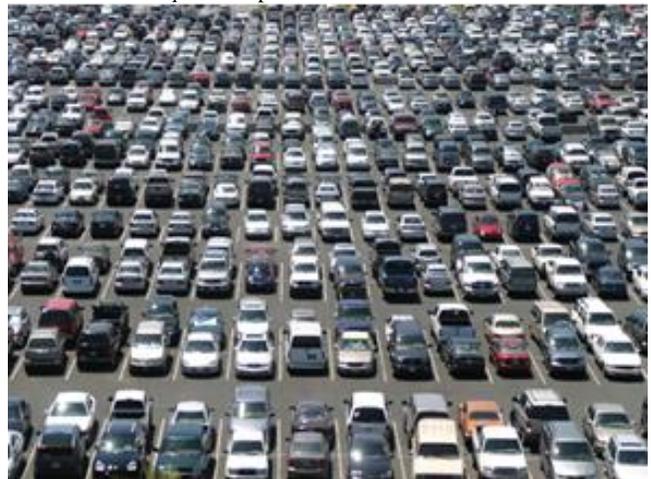

**Figure 1: Urban Sprawl affecting Parking Lots**

Urban sprawl can be defined as a pattern of urban and metropolitan growth that reflects low density, automobile-dependent, exclusionary new development and the fringe of settled areas often surrounding a deteriorating city. Among the traits of metropolitan growth, frequently associated with sprawl are unlimited outward extension of development, low density housing and commercial development, leapfrog development,

"edge cities,' and more recently "edgeless cities;" reliance on private automobiles for transportation, large fiscal disparities among municipalities, segregation of types of land use, race and class-based exclusionary housing and employment, congestion and environmental damage, and a declining sense of community among area residents [11]. The term urban sprawl generally has negative connotations due to the health, environmental and cultural issues associated with the phrase. Residents of sprawling neighborhoods tend to emit more pollution per person and suffer more traffic fatalities [2]. As a result people would start the trend of moving to neighborhood low density areas, and to meet their requirements, more houses, parking lots (see Figure 1), roads (see Figure 2), shops etc. should be constructed which gradually leads to expansion of sprawl to those areas as well.

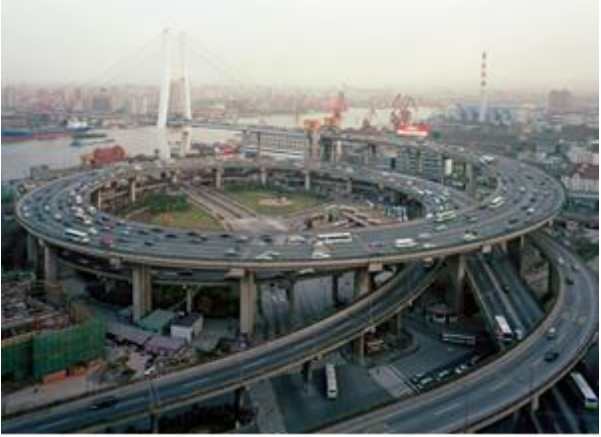

**Figure 2: Urban Sprawl affecting Roads**

In our work, we aim to find the inherent relations between the variables related to sprawl. Although the relations and patterns explained in this paper are specific to New York, these can be generalized to simulate other urban areas in United States. The basic structure of all cities is almost the same. They have similar infrastructure design, socio-economic conditions, transportation and facilities. It is this similarity that has encouraged us to devise a method which will help decision-makers/engineers to identify sprawl conditions in their respective areas and thus help them design their cities accordingly in order to eliminate the conditions of sprawl. In this context we use geographic information system (GIS) data for analysis on spatiotemporal parameters.

A geographic information system integrates hardware, software and data for capturing, managing, analyzing, and displaying all forms of geographically referenced information. The power of a GIS comes from its ability to relate different pieces of information in a spatial context and to reach conclusions about pertinent relationships [17]. Most of the information we have about the world contains a location reference, placing that information at some point on the globe. For example, when rainfall information is collected, it is important to know where the rainfall occurs. This is done by using a location reference system, such as longitude and latitude, and perhaps elevation. Comparing the rainfall information with other information, such as the location of marshes across the landscape, may show that certain marshes receive little rainfall [5]. This fact may indicate that these marshes are likely to dry up, and this inference can help us make the most appropriate decisions about how humans should interact with the marsh. A GIS, therefore, can reveal important new information that leads to better decision-making [14].

The focus of this paper is to understand and identify urban sprawl by mining such geospatial data coming from a GIS. The results of this mining would thus help urban planners make better decisions by providing models to predict urban sprawl that guide the planning to make the areas sustainable. In addition, we are interested in finding out how some variables pertaining to sprawl affect others, e.g., how many trucks are used for transportation within a sampled place given a certain number of housing units. To the best of our knowledge, a model to predict urban sprawl and study its parameters does not exist in the literature so far. We thus address this problem in our work.

More specifically, the problem addressed in this paper is defined with two specific goals as follows:
*1. Predict the occurrence of urban sprawl based on given parameters*
*2. Analyze the impact of the parameters on each other and on the urban sprawl*

The broader impact of this work addresses urban sustainability as explained below. Urban sustainability involves a reexamination of urban development, including environmental, social and economic policies, politics and practices, and an acknowledgement of the role of cities in global environmental change [15]. Sustainable development implies improving the quality of life of a population within the capacity of Earth's finite resources. The needs of the present generation must be met, particularly those of the poor, without compromising the ability of future generations to meet their own needs. This is a dynamic process whereby the decision-makers involved in any area plan, implement and then re-examine their ideas and policies over time. In cities the goal of sustainability has been increasingly highlighted over the past few decades as problems and issues arise from unsustainable practices and developments [18]. Thus a spatial decision support system (SDSS) that predicts the occurrence of urban sprawl by mining data on geographic information systems, and helps understand the relationships between parameters affecting urban sprawl, would have a positive impact on urban sustainability. It would help urban planners, city dwellers and other related personnel make better decisions that help to promote sustainable growth and development.

The rest of this paper is organized as follows. Section 2 describes our proposed approach to address the problem of urban sprawl. Section 3 describes the implementation and evaluation of the SDSS developed to predict urban sprawl. Section 4 outlines related work. Section 5 gives the conclusions and future work.

## 2. PROPOSED APPROACH

A spatial decision support system (SDSS) is proposed herewith as an approach to address the problem of predicting urban sprawl and analyzing the impact of its parameters. A SDSS is an intelligent, interactive computer-based system designed to assist in decision-making while solving a semi-structured spatial problem. It is designed to assist the spatial planner with guidance in making land use decisions. This entails use of spatiotemporal databases to store and process the geospatial data, a library of potential models that can be used to forecast the possible outcomes of decisions, and an interface to aid the user's interaction with the computer system and to assist in analysis of outcomes [20]. In this paper, we study the effects of the interesting variables and predict the occurrence of urban sprawl. We also study the impacts of the concerned variables on each other. In order to discover the patterns and trends among these variables, two classical data mining algorithms are deployed: Apriori for association rule mining [1] and J4.8 for decision tree classification [6]. These are adapted with specific reference to geospatial data.

The justification for decision tree classification in this work is that the primary goal of this work is the prediction of the discrete target attribute 'sprawl' based on other relevant attributes. Thus decision trees would be well-suited based on their paths that represent the concerned attributes along with their values, and the leaves that represent the decisions, i.e., predicting the occurrence of sprawl.

The justification for association rule mining is that the secondary goal here is to understand the correlation between the attributes pertaining to sprawl. Discovering patterns in the form of association rules of the type A=>B would be very useful here to indicate how one attribute impacts another.

We now describe the steps of our proposed approach involved in building the SDSS. There are three main steps as listed below.
- Data Preprocessing Step
- Association Rule Mining Step
- Decision Tree Classification Step.

In order to explain the concepts in these respective steps, we use a running example based on the state of New York. Real GIS data from New York counties is used in our work. Counties in NY are studied to distinguish areas affected by urban sprawl. It is to be noted that although NY is famous for its metros, it still has counties with a strong rural character. We thus use it as an exemplary data set.

## 2.1 Data Preprocessing Step

In order to discover knowledge from the GIS data, we first map it by superimposing a shape file on the urban land use file and then extract relevant information for conversion to suitable formats as appropriate for the concerned mining algorithms. For this purpose we use the ArcGIS software briefly described below.

ArcGIS is a GIS software package of Economic and Social Research Institute (ESRI), which is used for working with maps and geographic information [19]. It is used for:

- Creating and using maps
- Compiling geographic data
- Analyzing mapped information
- Sharing and discovering geographic information
- Using maps and geographic information in a range of applications
- Managing geographic information in a database.

Using ArcGIS data preprocessing is conducted as follows, referring to the example New York data set. This NY dataset consisting of 27 variables, excluding shape file, are all continuous data. A partial snapshot of this data appears in Figure 3.

These variables consist of demographics, socio-economic conditions, infrastructure usage, and accidental reports. The dataset consists of data for the 62 counties for the years 2000 and 2010. Among these 27 variables, the last one is the target attribute which defines the presence of sprawl occurrence. This target attribute is finalized based on a project about "Measuring the health effects of sprawl" done by smart growth America [10]. The collected dataset for the years 2000 and 2010, with a county based shape file for New York, as shown in Figure 4, is plotted as two separate maps using the Arc GIS software to show the urban sprawl in both the years. The map representation gives a clear visual depiction of the spatial data. The steps followed to plot the map are listed below with reference to the given example.

- Arc GIS 10.1 is opened to add the shapefile of NY.
- Under properties, counties are named using "Label", selecting 'Name', from the attribute table.
- Using the Join option the data file is joined with the existing attribute table of New York state.
- Two separate maps displaying Urban Sprawl in year 2000 and year 2010 are prepared.

**Figure 3: Partial Snapshot of NY State Urban Land Use Data**

**Figure 4: Shape File for New York State**

These maps prepared using ArcGIS are shown in Figures 5 and 6 respectively. In these maps, the counties shown in red indicate those with sprawl and the ones is green represent the absence of sprawl or low-intensity sprawl. By comparing these figures we can see that 5 more counties were affected with urban sprawl in 2010 than in 2000. In the map for 2000 the counties Putnam, Orange, Ditches are not in the list of sprawl-affected ones. But the expansion of the nearby highly dense counties affects these counties over those 10 years.

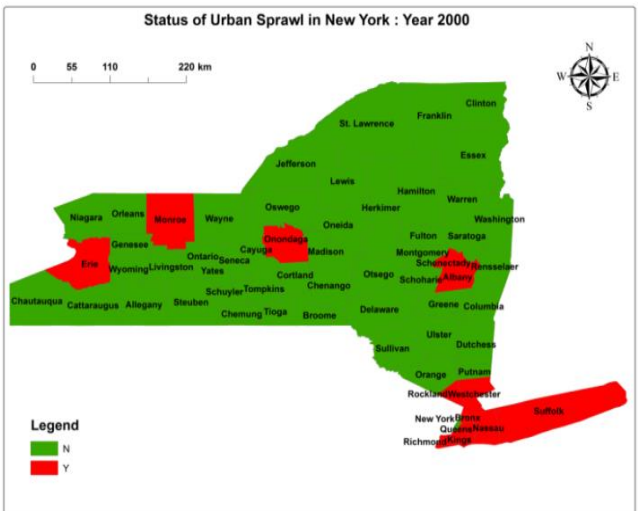

**Figure 5: NY Map with Presence of Sprawl for Year 2000**

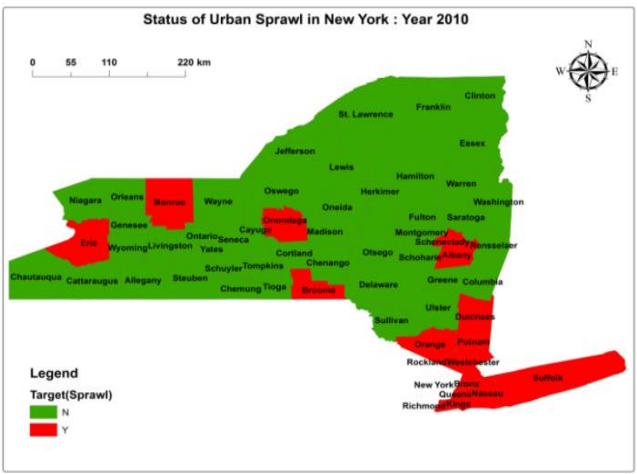

**Figure 6: NY Map with Presence of Sprawl for Year 2010**

The data preprocessing using ArcGIS sets the stage for discovering knowledge from the GIS data through the data mining techniques of association rules and decision trees as described in the next two subsections.

## 2.2 Association Rule Mining Step

The goal of the association rule mining step is to understand the causal relationships between pertinent variables in the geospatial data. We adapt the classical Apriori algorithm [1] for association rule mining in order to discover relationships among the various spatial attributes affecting urban sprawl.

Although this is the secondary goal of our work, we conduct this prior to decision tree classification. This is in order to gain an understanding of individual parameter impacts on each other before predicting how they all impact urban sprawl. Since the GIS data is spatial, we write a Java program based on the Apriori algorithm adapted to geospatial data. We carry out further processing within the algorithm such that continuous data are mapped to binary attributes, implementing relevant filters as needed. After running our implementation of Apriori for spatial data over the NY dataset, we get some association rules, examples which are shown in Figure 7.

[BirthRate_Range2 -> GasolineStations_Range2 Target_Sprawl,
Income_Range3 -> HousingUnits_Range3 Target_Sprawl,
Asians_Range3 -> HousingUnits_Range3 Target_Sprawl,
HousingUnits_Range3 -> BirthRate_Range2 Target_Sprawl,
Target_Sprawl -> BirthRate_Range2 GasolineStations_Range2,
GasolineStations_Range2 -> BirthRate_Range2 Target_Sprawl,
BirthRate_Range2 Target_Sprawl -> Income_Range3 HousingUnits_Range3, Income_Range3 Target_Sprawl -> BirthRate_Range2 HousingUnits_Range3, Income_Range3 BirthRate_Range2 -> HousingUnits_Range3 Target_Sprawl,
Asians_Range3 Target_Sprawl -> BirthRate_Range2 HousingUnits_Range3, Asians_Range3 BirthRate_Range2 -> HousingUnits_Range3 Target_Sprawl, Asians_Range3 Income_Range3 -> HousingUnits_Range3 Target_Sprawl,
HousingUnits_Range3 Target_Sprawl -> Income_Range3 BirthRate_Range2, Income_Range3 HousingUnits_Range3 -> BirthRate_Range2 Target_Sprawl, Asians_Range3 HousingUnits_Range3 -> BirthRate_Range2 Target_Sprawl,
BirthRate_Range2 HousingUnits_Range3 -> Income_Range3 Target_Sprawl]

**Figure 7: Examples of Association Rules over GIS Data**

The ranges shown in this figure are explained as follows. In order to apply data mining algorithms efficiently, continuous variables are converted into discrete variables by grouping them into various ranges. Depending on the values of the variables they are grouped into ranges numbered 2, 3 and 4 respectively. For example, in Figure 7, an association rule states that if the income rate is in Range 3 of income (greater than 10 per 1000 population) then the housing units in that area would be in Range 3 of housing (greater than 100000). If these conditions prevail, it would lead to urban sprawl.

Having understood the impact of these parameters on urban sprawl, through association rule mining, we now proceed with decision tree classification to predict the occurrence of sprawl.

## 2.3 Decision Tree Classification Step

The purpose of the decision tree classification step is to achieve the most important goal in our work, namely, prediction of whether urban sprawl occurs given certain conditions. We deploy the popular J4.8 algorithm for decision tree classification [6] over the spatial GIS data. Adaption is performed similar to the Apriori algorithm. Figure 8 shows a partial snapshot of the decision trees learned from the given GIS data. Following the example of the NY data set, consider the 27 attributes. Since population density and percentage of Asians have most impact on sprawl and had comparatively very high impact when compared with other variables, we show those patterns in the J4.8 results in this figure. The ranges in the figure are similar to those in association rule mining, depicting the respective categories of the parameters.

Additionally, in this step, bagging and boosting [7] are also carried out. Bagging, which generates repeated bootstrap samples of the data, is useful in order to learn a more generic hypothesis, in this case, the knowledge discovered from the GIS data to predict the target attribute "sprawl" based on ranges. Boosting, which adjusts the weights of the training instances is useful to incorporate their relative importance to enhance the discovered knowledge. Thus, bagging and boosting are conducted in the development of our SDSS in order to improve the prediction rules with the intention of enhancing performance in the SDSS for prediction. Results after bagging and boosting appear in Figures 9 and 10 respectively.

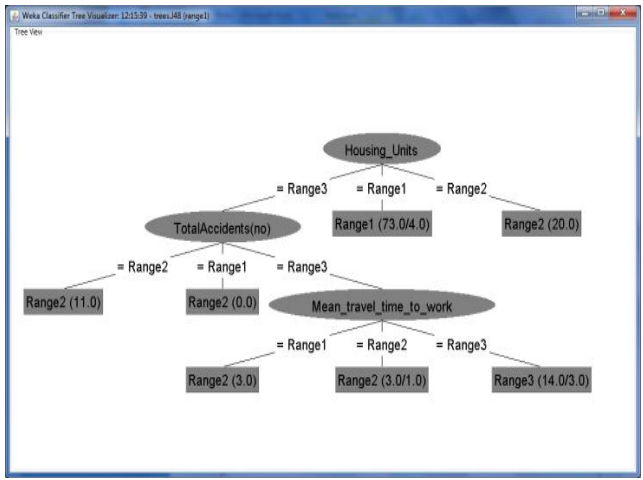

**Figure 8: Partial Snapshot of Output from J4.8 over GIS Data**

```
REPTree
============
TotalPersonaIncome< 11713160
|   Employed(of tot pop) < 18.88 : Y (2/0) [1/0]
|   Employed(of tot pop) >= 18.88 : N (61/1) [30/0]
TotalPersonaIncome>= 11713160 : Y (19/0) [11/0]
Size of the tree : 5
REPTree
============
TotalPersonaIncome< 11286795 : N (65/3) [33/1]
TotalPersonaIncome>= 11286795 : Y (17/1) [9/0]
Size of the tree : 3
REPTree
============
White people(of tot pop) < 82.31 : Y (20/2) [10/1]
White people(of tot pop) >= 82.31 : N (62/0) [32/1]
Size of the tree : 3
REPTree
============
Percentage of foreign born < 5.25 : N (56/0) [27/1]
Percentage of foreign born >= 5.25
|   FarmLand(Acres) < 44.7
|   |   TotalPersonaIncome< 373943200 : Y (15/0) [5/0]
|   |   TotalPersonaIncome>= 373943200 : N (2/0) [1/0]
|   FarmLand(Acres) >= 44.7 : N (9/2) [9/3]
Size of the tree : 7
REPTree
============
TotalPersonaIncome< 10641129
|   FarmLand(Acres) < 71.6
|   |   Percentage of foreign born < 4.7 : N (6/0) [2/0]
|   |   Percentage of foreign born >= 4.7 : Y (6/2) [5/2]
|   FarmLand(Acres) >= 71.6 : N (51/0) [25/0]
TotalPersonaIncome>= 10641129 : Y (19/0) [10/1]
```

**Figure 9: Output after Bagging**

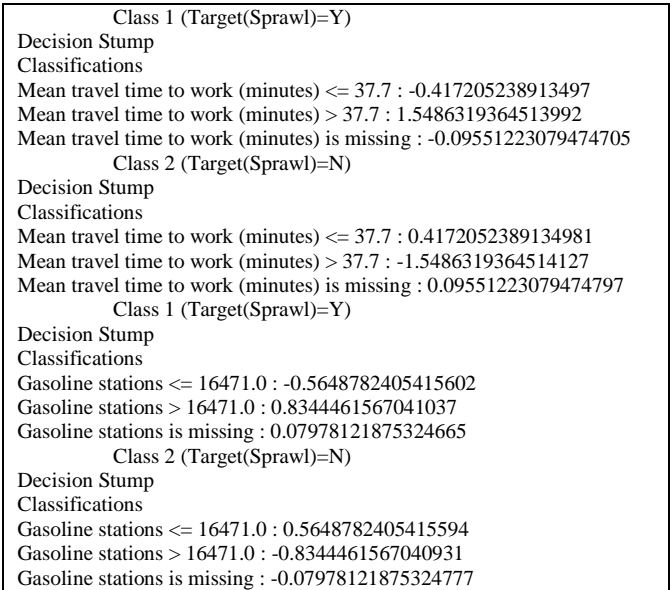

```
        Class 1 (Target(Sprawl)=Y)
Decision Stump
Classifications
Mean travel time to work (minutes) <= 37.7 : -0.417205238913497
Mean travel time to work (minutes) > 37.7 : 1.5486319364513992
Mean travel time to work (minutes) is missing : -0.09551223079474705
        Class 2 (Target(Sprawl)=N)
Decision Stump
Classifications
Mean travel time to work (minutes) <= 37.7 : 0.4172052389134981
Mean travel time to work (minutes) > 37.7 : -1.5486319364514127
Mean travel time to work (minutes) is missing : 0.09551223079474797
        Class 1 (Target(Sprawl)=Y)
Decision Stump
Classifications
Gasoline stations <= 16471.0 : -0.5648782405415602
Gasoline stations > 16471.0 : 0.8344461567041037
Gasoline stations is missing : 0.07978121875324665
        Class 2 (Target(Sprawl)=N)
Decision Stump
Classifications
Gasoline stations <= 16471.0 : 0.5648782405415594
Gasoline stations > 16471.0 : -0.8344461567040931
Gasoline stations is missing : -0.07978121875324777
```

**Figure 10: Output after Boosting**

## 3. IMPLEMENTATION & EVALUATION

After discovering knowledge from the spatial GIS data, the results of the association rule mining and decision tree classification are used to implement the SDSS. The most important parameters affecting urban sprawl as obtained from the knowledge discovery process are critical in the SDSS development and are used to design questions that targeted users could pose to the system.

For example, in the NY state data, we find that population density is the major cause of urban sprawl. Moreover, due to the increase in population other related factors also increase. We also find interesting results such as how the number of housing units, truck transportations and gas stations relate to each other in sprawl-affected areas. Based on this learning, the anticipated questions in the SDSS are designed which guide user interaction. These cater to various user queries, a few examples of which are listed below.

- If the unemployment rate is greater than a certain value, would sprawl be likely to occur?
- If personal income is in a given range & percentage of Asians is above a certain value, would sprawl occur?
- What is the relationship between birth rate and sprawl occurrence?
- What is the relationship between housing units and gas stations?

Note that the first query predicts the likelihood of sprawl based on a single parameter while the second one predicts its likelihood based on multiple parameters. The third one is to understand the impact of a given parameter on sprawl whereas the fourth one is to understand the impact of parameters on each other. Such queries are designed during the SDSS implementation based on the GIS data mining and serve as the basis for creating the user interface for interaction with the system.

This SDSS is implemented in Java using the knowledge discovered by mining the GIS data. A screen-dump of the SDSS is shown in Figure 11. This system offers an interface through which the user can pose queries in the form of several decision-making scenarios to the system.

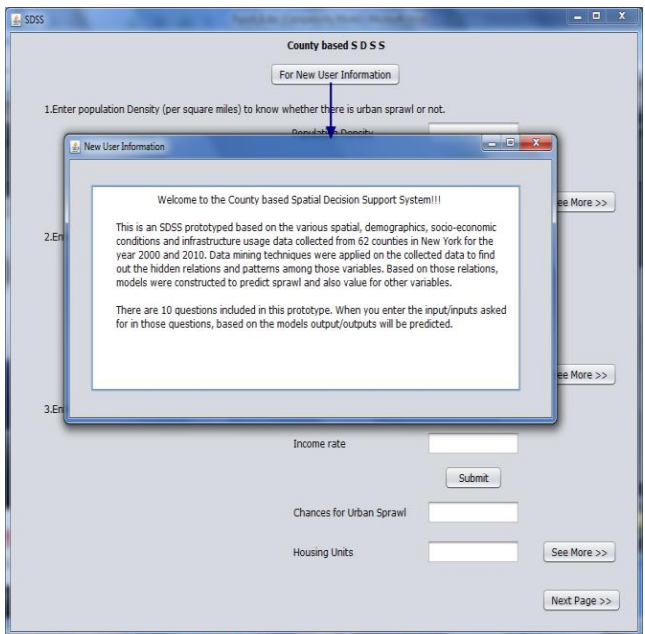

**Figure 11: Prototype SDSS for Urban Sprawl**

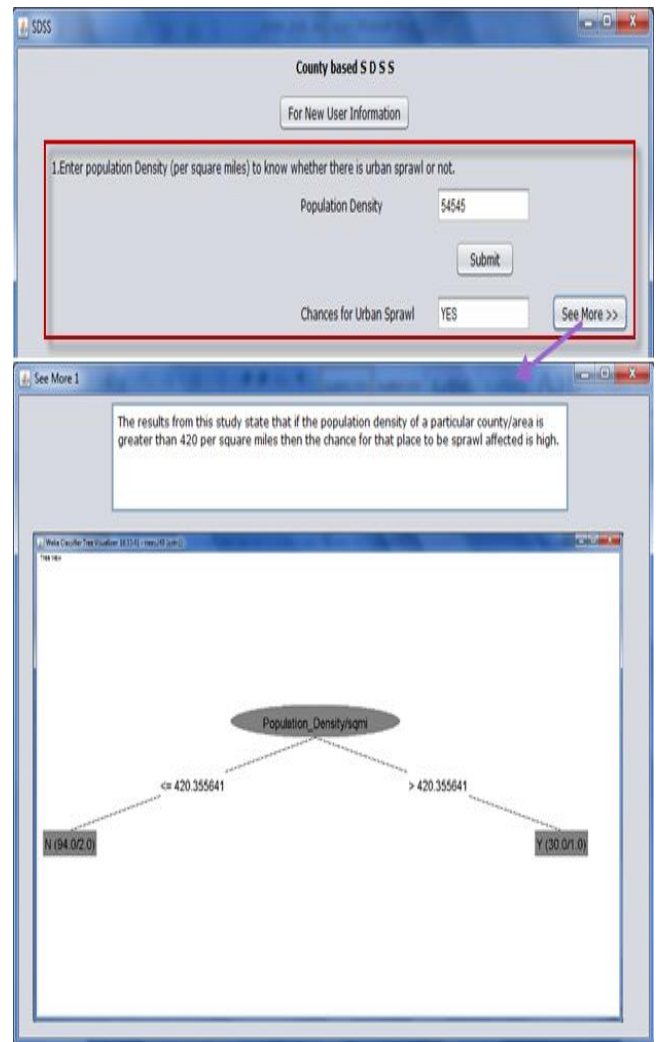

**Figure 12: How Population Density affects Sprawl**

We conduct a thorough evaluation of this SDSS by running various user queries. Two such sample queries are shown in the scenarios below, along with the responses of the SDSS to the respective scenarios.

*Scenario 1:* The user queries whether sprawl occurs by entering a certain population density (54545 as in Figure 12). The system predicts that sprawl occurs in this case (YES). In order to understand why system gives this response, the user can select the "See More" option of the SDSS. It shows that based on decision tree learning over the given data set, if population density per square miles is more than 420, then sprawl is likely to occur. This example predicts the occurrence of sprawl shows the impact of a given parameter on the target attribute "sprawl".

*Scenario 2:* The user needs to know how many houses would need electricity for heating and cooling, given the total number of housing units at that area (76767 as in Figure 13). The SDSS responds "less than 20,000" as shown. The "See More" option indicates how this was learned by association rules. This example shows the impact of sprawl-related parameters on each other.

Likewise, users can pose many queries to the SDSS to predict sprawl occurrence based on given conditions and to understand the impact of sprawl-causing factors on each other. This prototype SDSS is useful to urban planners and geoscientists and sets the stage for the development of a large scale SDSS for predictive analysis in urban planning by mining over big data with the potential use of cloud technology and other advances. More algorithms and solutions would be proposed in the further development of the SDSS in the future.

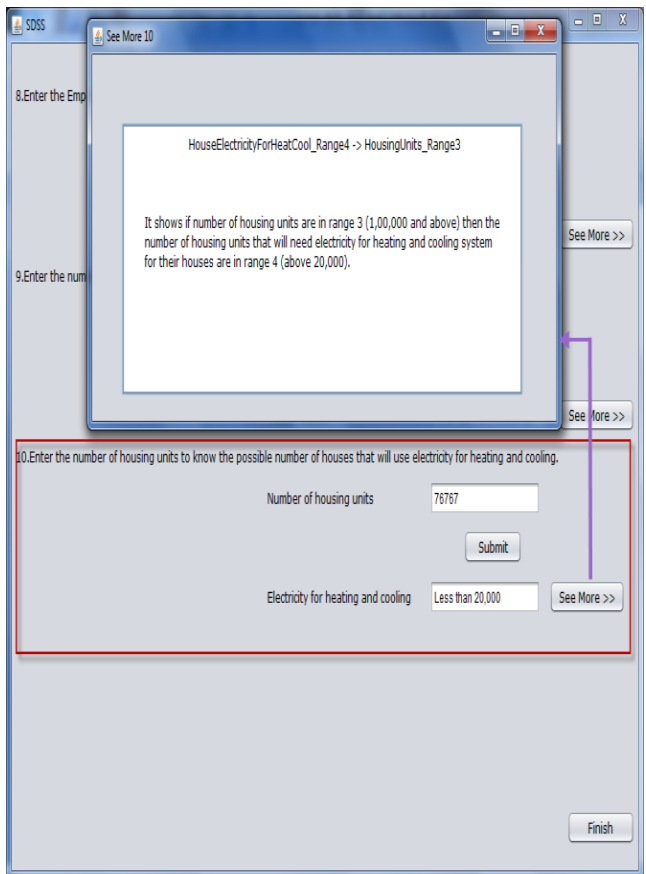

**Figure 13: Relation between Housing Units & Possible Number of Houses with Electricity for Heating, Cooling**

## 4. RELATED WORK

In recent years geospatial technology is developing rapidly. New techniques in the GIS field and large resources in the form of digital imagery, statistical data are encouraging more and more researchers to put in efforts into topics such as urban sprawl and sustainability, which is now a big concern among geologists, environmentalists and city planners. We cite some of the literature herewith.

In Jiang et al. [3], 2007 they take Beijing as a case study and put forward that urban sprawl can be measured from spatial configuration, urban growth efficiency and external impacts, and then develops a geo-spatial indices system for measuring sprawl, using a total of 13 indicators. Thus, the authors try to measure the sprawl or rate the sprawl after it occur. However, in our work, we predict the occurrence of sprawl and also find the patterns that caused the sprawl.

Luo et al. [4], strive to model urban expansion with the use of geographic information systems by remote sensing. The focus of their work is more on building a model through remote sensing technologies. Our work assumes that the GIS data has already been collected and works further to analyze that data to predict the occurrence of sprawl in the future and study the impact of the parameters causing the sprawl.

In the study conducted by Sudhira et al. [12], the authors study sprawl using spatial data along with other attributes. The study area is India, and their study attempts to identify sprawl, quantify it by defining new metrics, understand the dynamic process and subsequently model the same. The authors in this study have used statistical analysis. However, we consider data mining algorithms over GIS data to find the sprawl causing patterns.

The authors of Sun et al. [13], implement land use analysis for the City of Calgary, Alberta, Canada, using an object-oriented approach. They aim to simulate the land use pattern using Markov Chain analysis and Cellular Automata analysis based on the interactions between the land use and the transportation network. However, they do not develop tools to predict sprawl occurrence to assist urban planning. In our work we analyze GIS data with the adaption of data mining techniques over the geospatial attributes. Also, we predict the manner in which land use is likely to develop in the future. Moreover, we develop a prototype SDSS to assist decision-making in urban planning by predicting urban sprawl and its parameter impact.

## 5. CONCLUSIONS

We have addressed the issue of predicting urban sprawl by mining GIS data. We have collected data for variables directly or indirectly related to urban sprawl and adapted the data mining algorithms of Apriori for association rule mining and J4.8 for decision tree classification to this spatiotemporal data. Based on the results of knowledge discovery over this data, we have implemented a prototype SDSS that can assist in decision-making on urban sprawl.

This can be used by urban planners, city dwellers and any other users interested in finding out the likelihood of sprawl occurrence and understanding how sprawl-causing variables affect each other. This work on the whole contributes to the computer science community of spatiotemporal data mining and the geo-science community of urban sustainable development. It spans the realm of geospatial intelligence has an impact on the environment and public health.

Our future work includes deploying this prototype as the basis for research and development of a large scale SDSS in urban planning by knowledge discovery over big data. This could potentially entail the proposal of advanced techniques in cloud computing, and / or new paradigms in areas overlapping GIS and big data to solve the concerned problems. Its broader impact would be to enhance urban sustainability by helping to make decisions for promoting the development of sustainable cities and surrounding areas.